\documentclass[conference]{IEEEtran}
\pdfoutput=1
\usepackage{amsmath,amssymb,amsthm}
\usepackage{graphicx}
\usepackage{hyperref}
\usepackage[table]{xcolor}
\usepackage{booktabs}
\usepackage{subcaption}
\usepackage{cleveref}

\theoremstyle{plain}

\theoremstyle{definition}

\theoremstyle{remark}

\begin{document}

\title{An Entropic Metric for Measuring Calibration of Machine Learning Models}

\author{
\IEEEauthorblockN{Daniel James Sumler\IEEEauthorrefmark{1}, Lee Devlin\IEEEauthorrefmark{1}, Simon Maskell\IEEEauthorrefmark{1}, and Richard O. Lane\IEEEauthorrefmark{2}}
\IEEEauthorblockA{\IEEEauthorrefmark{1}The University of Liverpool, UK.}
\IEEEauthorblockA{\IEEEauthorrefmark{2}QinetiQ, Great Malvern, UK.}
}

\maketitle

\begin{abstract}
Understanding the confidence with which a machine learning model classifies an input datum is an important, and perhaps under-investigated, concept. In this paper, we propose a new calibration metric, the Entropic Calibration Difference (ECD). Based on existing research in the field of state estimation, specifically target tracking (TT), we show how ECD may be applied to binary classification machine learning models. We describe the relative importance of under- and over-confidence and how they are not conflated in the TT literature. Indeed, our metric distinguishes under- from over-confidence. We consider this important given that algorithms that are under-confident are likely to be “safer” than algorithms that are over-confident, albeit at the expense of also being over-cautious and so statistically inefficient. We demonstrate how this new metric performs on real and simulated data and compare with other metrics for machine learning model probability calibration, including the Expected Calibration Error (ECE) and its signed counterpart, the Expected Signed Calibration Error (ESCE).
\end{abstract}

\section{Introduction}
\label{sec:introduction}
Calibration of probabilities is an important and often-overlooked concept when developing machine learning (ML) models. Usually, accuracy is the main metric used to calculate how well an ML model performs in terms of predicting a class for unseen data. Generally speaking, the closer the accuracy is to 100\%, the better the model is deemed to be. However, this does not take into account the probability of predictions that the model outputs, which can be just as important, if not more, than the accuracy. 

In binary classification, a probability greater than a threshold, typically 0.5, is enough to decide whether an input belongs to one of two classes. While accuracy informs whether a classification is correct, a calibration metric informs how well the confidence probabilities match the true proportions of correct decisions. For example, a model that always outputs a probability of 0.6 for class label 1, but always gets this classification correct, should produce a poor calibration score, as even though the model has classified the output correctly, it has low confidence in that decision.

Calibration has become even more important as of late, as the research of Guo et al. \cite{guo2017calibration} reveals that while modern neural networks are more accurate than ever, they are also badly calibrated. This could be attributed to the over-confidence of said networks due to the large amount of data they are able to be trained on. 

A well-calibrated model can be defined as one that outputs probabilities that are representative of the real-life occurrences from the unseen data. For example if, on average, 70\% of people are correctly predicted to contract a certain disease, then one would expect the average probability outputted by a diagnosis model to be 0.7. A mathematical representation of calibration can be seen in (\ref{eq:callibraion}), where $x \in \{0,1\}$ is the true label, $y \in \mathbb{R}^d$ is an observed data sample of dimension $d$ belonging to a binary class $ k \in \{0,1\} $, and $p_k$ is the confidence in class $k$, while $\mathbb{P}$ is the true probability.

\begin{equation}
      \label{eq:callibraion}
     \mathbb{P} (x = k | y) = p_k
\end{equation}

In this paper, we present a novel calibration metric that addresses some weaknesses of some of the most commonly-used existing metrics. In section \ref{sec:lit_review}, we discuss existing calibration metrics that are widely discussed throughout the literature. In section \ref{sec:motivation}, we detail our motivations for ``safe'' calibration and why we feel our metric is necessary. Section \ref{sec:ecd} explains our new metric and details how the results can be interpreted. In section \ref{sec:results}, we test our metric on simulated and real data, and compare it with other popular calibration metrics. Finally, section \ref{sec:conclusion} concludes the paper. 

\medskip

\section{Existing metrics for model calibration}
\label{sec:lit_review}

In this section, we explore the literature and highlight some important calibration metrics, from early pioneers of the field to the widely-used modern standards. 

There exist various different methods that attempt to solve the problem of badly calibrated models. Some attempt to do this by altering their training process, such as using a loss function that tries to address neural network calibration as it is being trained \cite{neo2024maxent}, while other methods attempt to change a model's output probabilities (\cite{Boken2020-kb, guo2017calibration}). The former method is useful if we want to train a model; however, this is not always the case. Occasionally, pre-trained models will need their existing probabilities calibrated, which is where the latter method is used. However, before calibrating a model's output probabilities, we must first find out whether the model is already calibrated or not and to what extent. For this, we use calibration metrics, and the remainder of this section defines and discusses their benefits and drawbacks. 

\subsection{Brier Score}

One of the most influential examples of a calibration metric is the Brier Score, originally proposed for use with weather forecast calibration \cite{brier1950verification}. However, it has lost some popularity due to the emergence of metrics that address its disadvantages \cite{spiegelhalter1986probabilistic}.

The Brier Score, as seen in (\ref{eq:brier}), is a weighted version of the Mean Squared Error (MSE) equation which incorporates a predicted probability, $f_t$, and the actual outcome, $o_t$. 
\begin{equation}
    \label{eq:brier}
    BS = \frac{1}{N}\sum^N_{t=1}\left(f_t - o_t\right)^2
\end{equation}
The Brier score computation outputs a normalised value between 0 and 1, with the former showing perfect calibration and the latter depicting complete miscalibration. Due to the nature of the metric, the Brier Score incorporates both accuracy and calibration. It prioritises models that are confident over models that may be well calibrated. Incorrect predictions may still result in a good Brier Score as long as the majority of the other predictions are confident and accurate \cite{brier_score_bad, huang2020preservation}. 

\subsection{Reliability Diagram}
While the Brier Score gives a numerical value to indicate the level of calibration a model has, other methods have been developed to allow one to visually assess the calibration of a model. The most popular method for doing this is to use reliability diagrams. First brought to the limelight by DeGroot and Fienberg \cite{degroot1983comparison}, and then further expanded upon by Niculescu-Mizil and Caruana \cite{niculescu2005obtaining}, reliability diagrams are a widely-used method to visually display calibration. 

This visual technique works by partitioning the probabilities that a model outputs into a number of bins, where each bin represents a probability interval. It is possible to use any number of bins, but one must keep in mind the bias-variance trade-off as highlighted by Nixon et al. \cite{Nixon_2019_CVPR_Workshops}. They specify that when the number of bins is increased, this results in a lower population per bin, and, in turn, a larger variance. This will, however, also result in a lower bias. Therefore, the number of bins should be fine-tuned for each problem.

\begin{figure}[ht]
    \vskip 0.2in
    \begin{center}
        \centerline{\includegraphics[width=\columnwidth]{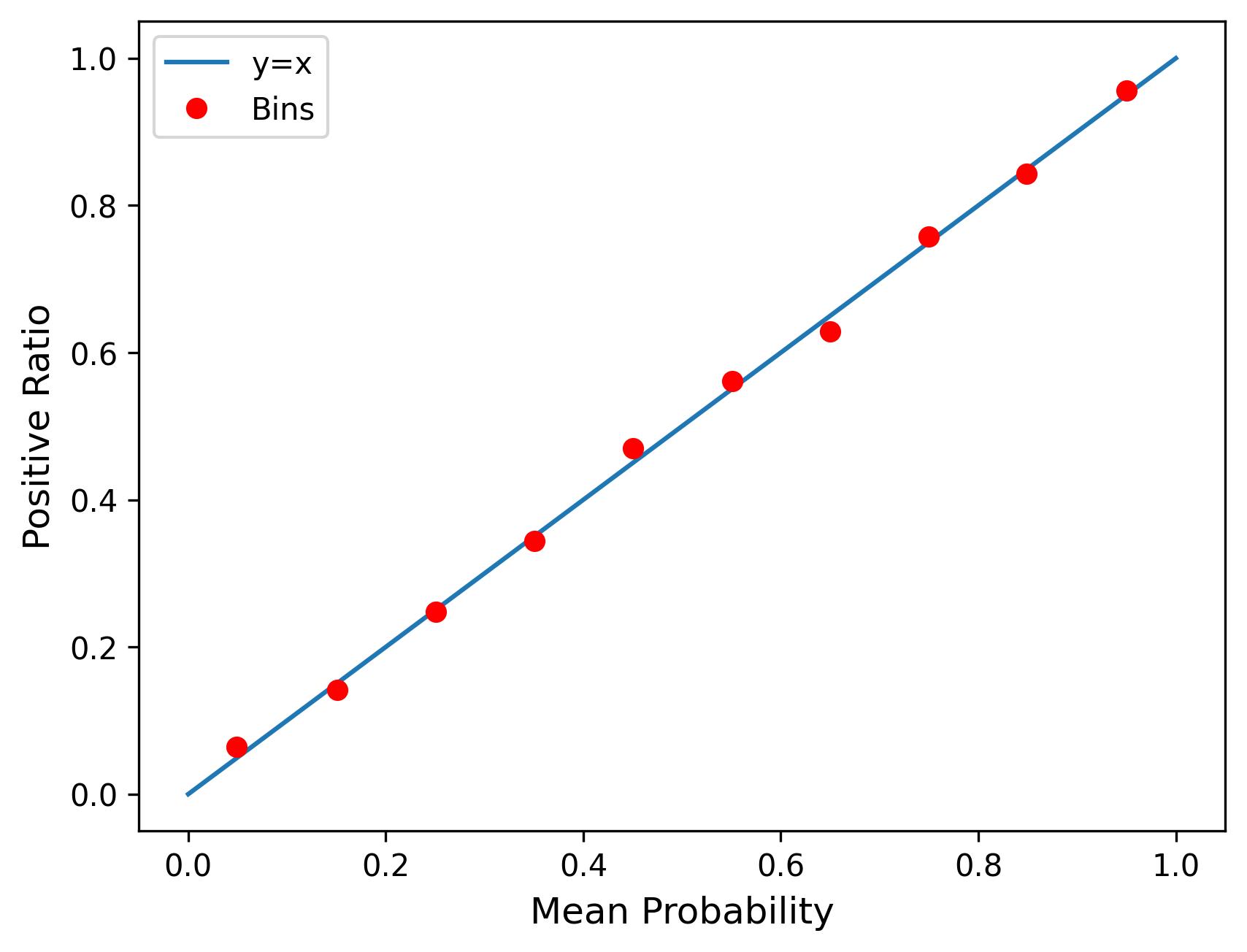}}
        \caption{A reliability diagram for a well-calibrated model. Blue line represents $ratio = probability$, red dots represent bins.}
        \label{fig:rel_diagram}
        \end{center}
    \vskip -0.2in
\end{figure}

Once every probability has been assigned to the appropriate bin, the average predicted probability and fraction of positive labels can be calculated for each bin. The fraction of positives is then plotted against the average predicted probability, with a reference line $fraction = probability$ that shows what a perfectly calibrated model would look like. Any points above the line are under-confident, while any points below it are over-confident. Figure \ref{fig:rel_diagram} shows what well-calibrated probabilities look like. 

Reliability diagrams can be a good tool for visualising a model's calibration. However, they do not, in themselves, return a calibration score, and the diagram can be misleading when some bins are sparsely populated.

\subsection{Expected Calibration Error}

Initially proposed by Naeini et al. \cite{naeini2015obtaining}, the Expected Calibration Error (ECE) is one of the most widely used calibration metrics. It is a simple, yet effective, formula which uses the same method of binning as reliability diagrams, but produces a normalised calibration score between 0 and 1, similar to the Brier Score. 

ECE works by calculating the accuracy and confidence of each bin in a reliability diagram. The accuracy is the traditional definition, where the number of correct predictions is divided by the total number of predictions in the bin. The confidence refers to the average probability in a bin. It is not explicitly stated in the original ECE paper, but Guo et al. \cite{guo2017calibration} use the maximum class probability to calculate the confidence of a model output in multiclass classification scenarios. This means that, for a multiclass classifier, the minimum probability in all bins will be 0.5, and half of the bins will be unpopulated. In binary classification instances, the estimated probability, $\hat{p_i}$, of the positive class is used when calculating the mean confidence for bin $B_m$, as seen in (\ref{eq:conf}), and the ratio of positives is taken instead of the accuracy (\cite{Silva_Filho2023-lw, Guilbert2024-tk}), as shown in (\ref{eq:fracpos}), where $y_i$ is the true label. These are the same values used in the reliability diagram. The ECE of each bin is then calculated by obtaining the absolute difference between (\ref{eq:fracpos}) and (\ref{eq:conf}). This ECE value is weighted depending on how populated each bin is, as shown in (\ref{eq:ECE}), where $N$ is the total number of probabilities across all bins.

\begin{equation}
    \label{eq:conf}
    conf(B_m) = \frac{1}{|B_m|}\sum_{i = 1}^{|B_m|} \hat{p}_i
\end{equation}
\begin{equation}
    \label{eq:fracpos}
    frac_{pos}(B_m) = \frac{1}{|B_m|}\sum_{i = 1}^{|B_m|} 1(y_i = 1)
\end{equation}
\begin{equation}
    \label{eq:ECE}
    ECE = \sum^{M}_{m=1} \frac{|B_m|}{N} |frac_{pos}(B_m) - conf(B_m)|
\end{equation}
While ECE is widely-used in the literature, it has some shortcomings. One of the most noticeable is the use of bins, and the fact that the user needs to decide on an optimal binning strategy. Due to its positive fraction and confidence terms, as shown in (\ref{eq:fracpos}) and (\ref{eq:conf}), changing the number of bins can change the final ECE score. This also evokes the bias-variance trade-off mentioned previously \cite{Nixon_2019_CVPR_Workshops}. 

Additionally, the ECE equation deals with averages within bins rather than individual sample probabilities and their respective true labels. Due to this, outliers (such as a wildly incorrect prediction) may not have a great impact on the final calibration score. While it may be the case that this is a good thing, since a model should not be heavily penalised for a single mistake, it could be considered much more important in sensitive models, such as ones that predict the probability of a medical patient having a certain disease. 

\subsection{Expected Signed Calibration Error}

The Expected Signed Calibration Error (ESCE) was originally proposed by Verhaeghe et al. \cite{verhaeghe2023generalizable} for use with machine learning models detecting atrial fibrillation. As the name implies, this is a signed version of the ECE equation. The only difference with this metric is that the absolute value of the accuracy and confidence is no longer taken, resulting in a possible negative value. This expands the possible range of values from $[0, 1]$ to $[-1, 1]$. This, in turn, makes the metric more informative, as it is able to show under-confidence and over-confidence. Equation (\ref{eq:ESCE}) shows the ESCE formula. The equations to calculate the confidence and fraction of positives remain the same as shown in (\ref{eq:conf}) and (\ref{eq:fracpos}), respectively.
\begin{equation}
\label{eq:ESCE}
ESCE = \sum^M_{m=1} \frac{|B_m|}{N} (frac_{pos}(B_m) - conf(B_m))
\end{equation}
Results using ESCE are comparable to those for ECE, with the added benefit of being able to represent under- and over-confidence. The results which are presented by Verhaeghe et al. \cite{verhaeghe2023generalizable} show that there can be some cancellation between under- and over-confident bin values, resulting in a low ESCE score. Since it is not always evident whether a low ESCE score indicates good calibration or cancellation of bin values, it is unlikely that it would be used as a lone metric and would instead benefit more from being used alongside ECE.

The metrics explored above treat over- and under-confident predictions as equals. This works well for models that strictly require all probabilities to be well-calibrated; however, this is not always attainable. We address this problem in the following sections by presenting a different way of thinking about calibration.

\section{Motivation}
\label{sec:motivation}

In this section, we present some background knowledge that frames the metric we propose in this paper, and what we would like to call ``safe'' calibration. In the following subsections, we talk about the target tracking (TT) literature which is the inspiration for ``safe'' calibration. We also propose our metric, the Entropic Calibration Difference, and show how this fits into the TT field, and how it can be adapted for general use with ML model calibration.

\subsection{Target Tracking}
A fundamental goal of TT is to derive the state (e.g. position and velocity) of an object over time through noisy measurements. This is accomplished by using algorithms called target trackers.

Target tracking algorithms generally consist of multiple components; however, one of the core algorithms is track filtering. Commonly used methods include Kalman Filters and Particle Filters \cite{5512258}. Target trackers are often used to process 2D polar measurements. When the range is relatively well estimated, the probability distributions involved each resemble a banana. It transpires that the Taylor series used by the Extended Kalman Filter (EKF) \cite{einicke1999robust} fails to adequately represent the uncertainty caused by the curvature of the distribution. This can lead an EKF to diverge over time since it attributes excessive confidence to the output of processing previous data relative to a newly acquired datum. Techniques such as the Unscented Kalman Filter (UKF) (\cite{julieruhlmann, wandermerwe}) attempt to address this using a form of quasi monte-carlo integration. In the TT literature there is then a need to quantify the extent to which a technique consistently under-estimates the uncertainty.

One of the more popular metrics is the Normalised Estimation Error Squared (NEES), shown in (\ref{eq:nees}). Note that this equation assumes that $y$ is 1D. This works by calculating the ratio between the actual estimation error (being the difference between the predicted and true states), and the predicted error covariance matrix. 

\begin{equation}
    \label{eq:nees}
    NEES = \frac{1}{N} \sum^N_{i=1} \frac{(y_i - \hat{y_i})^2}{\sigma^2_i}
\end{equation}
NEES is a good consistency metric for single-target tracking. In this paper, we propose the Entropic Calibration Difference (ECD) metric, which is inspired by NEES. One of the main aspects that the metric borrows from NEES is the concept of ``safe'' calibration.

\subsection{``Safe'' Calibration}
The aim of ``safe'' calibration is to determine whether an ML model can be deemed ``safe'' to use. If a model is not well calibrated, it runs the risk of being over-confident in incorrect answers, or under-confident in the correct answers. We would like to bring the MTT way of thinking into ML, where over-confidence is considered much worse than uncertainty or under-confidence. The theory behind this is that we would much prefer a model be uncertain in the correct class rather than confidently choose an incorrect class. To this end, we penalise overconfidence more than under-confidence and uncertainty. ECD is able to determine whether a model is well-calibrated and can be interpreted as whether the model is ``safe'' to use. 

For example, if a binary classification model was determining whether it is safe for a plane to land, over-confidence in the incorrect class could potentially be fatal. Therefore, we would prefer to have under-confidence in the correct class or uncertainty, rather than random confident guessing. 
\section{Entropic Calibration Difference}
\label{sec:ecd}

\subsection{Background and Definition}

NEES is based on a chi-squared distribution and therefore outputs the degree of freedom associated with its inputs. It works by finding the difference between a true value, $y$, and the prediction, $\hat{y}$, and dividing this value by the uncertainty, $\sigma$. If the uncertainty value is not reasonably accurate, the final NEES value will not accurately reflect the consistency of the state estimator. Equation (\ref{eq:nees}) shows the NEES equation for a single dimension.

For a system with well-calibrated uncertainty in its output, a NEES score of 1, or $d$ if $y$ is more than 1D, is expected on average and shows perfect consistency, as it reflects a balance between the squared prediction error and the predicted uncertainty. Values greater than 1 show over-confidence, as the prediction error is large relative to the predicted uncertainty. This suggests that the model underestimated its uncertainty, which led it to be too confident in its predictions. This leads to a higher value of NEES compared to under-confidence. Finally, a value smaller than 1 shows under-confidence as the prediction error is small relative to the predicted uncertainty. This suggests that the model overestimated its uncertainty, making it overly cautious about its predictions.
The behaviour of being able to find over- and under-confident predictions using only a single equation is desirable for calibration, as it gives a greater context regarding the model's predictions. Treating over- and under-confidence differently would also give us an idea of whether a model is ``safely'' calibrated.

Our metric is applicable to both TT and ML calibration or any other probabilistic model and does not require any parameters other than the true label and prediction. Equation (\ref{eq:ecd}) shows the ECD metric.

\begin{equation}
    \label{eq:ecd}
    ECD=\frac{1}{N} \sum^N_{i=1} \left[ \int \log p(x|y_i)p(x|y_i)dx - \log p(x_i|y_i) \right]  
\end{equation}
\noindent where $N$ is the total number of data points, $y_i$ are the measured data for data point $i$, $p(x|y_i)$ is the algorithm's estimate of the probability density or probability mass of true state $x$ given the measurement, and $x_i$ are the known true states for a test set.

The first term in the summand in (\ref{eq:ecd}), containing the integral, is the negative entropy of the predicted probability distribution, or expected log likelihood, for a particular data point. This is used to represent under-confidence. The second term of the summand is the log likelihood, which is used to represent over-confidence. It should be noted that if the entropy term were zero, the overall expression would be the negative log likelihood (NLL), which is a commonly used metric to measure the calibration of classifiers. In general, ECD measures the difference between expected and actual log likelihoods. In this case, we have a metric that can produce negative scores for under-confident values, and positive scores for overconfident values. However, unlike other calibration metrics such as ECE and ESCE, under- and over-confidence are not treated the same, as the metric follows ``safe'' calibration scoring.

\subsection{Proving Relation with NEES}

NEES can be proven to be a special case of ECD by substituting the Gaussian distribution into both equations. Equation (\ref{eq:gaussian_pdf}) is the Gaussian formula, (\ref{eq:ecd_gaussian}) is the Gaussian ECD, and (\ref{eq:nees_gaussian}) is the Gaussian NEES.

\begin{equation}
    \label{eq:gaussian_pdf}
    p(x|y_i) = \frac{1}{\sqrt{(2\pi|C_i|)}} e^{-\frac{1}{2}(x - \mu_i)^T C_i^{-1} (x - \mu_i)}
\end{equation}

\begin{equation}
    \label{eq:ecd_gaussian}
    ECD = \frac{1}{N} \sum^N_{i=1} \left[\frac{1}{2} (x - \mu_i)^T C_i^{-1} (x - \mu_i) - \frac{d}{2} \right]
\end{equation}

\begin{equation}
    \label{eq:nees_gaussian}
    NEES = \frac{1}{N} \sum^N_{i=1} \left[(x - \mu_i)^T C_i^{-1} (x - \mu_i)\right]
\end{equation}
$C$ is the covariance matrix, $x$ is the state vector, $y_i$ is the observation, and $\mu$ is the mean of the Gaussian. NEES should be equal to dimensionality $d$ in (\ref{eq:ecd_gaussian}) for a system to be consistent and the ECD score would be 0. 

\subsection{ECD for Discrete Variables}
To apply the ECD formula to binary classification problems, it is necessary to allow it to be used with discrete variables. This is a simple change, as noted in (\ref{eq:ecd_discrete}).
\begin{multline}    
    \label{eq:ecd_discrete}
    ECD = \frac{1}{N}\sum^N_{i=1} \\
    \left[ 
    \left[ \sum^K_{k = 1} p(x=k|y_i)\log p(x=k|y_i) \right] - \log p(x_i|y_i) \right]
\end{multline}
In this modified equation, we predict the probability that $x$ is of class $k$ given a piece of observed data $y_i$, where $N$ and $K$ are the total amount of data points and number of classes, respectively. This allows for an easy transition to an ECD formula for binary calibration. By substituting (\ref{eq:p_substitute}) and (\ref{eq:log_substitute}) into (\ref{eq:ecd_discrete}), we create the binary classification ECD formula in (\ref{eq:ecd_binary}).
\begin{equation}
    \label{eq:p_substitute}
    \hat{p_i} = p(x=1|y_i)
\end{equation}
\begin{equation}
\label{eq:log_substitute}
    \log p(x_i|y_i) = x_i \log\hat{p_i} + (1-x_i)\log(1-\hat{p_i})
\end{equation}
\begin{equation}
    \label{eq:ecd_binary}
    ECD = \frac{1}{N}\sum^N_{i=1}(\hat{p_i} - x_i)\log\left[\frac{\hat{p_i}}{1-\hat{p_i}} \right]
\end{equation}
With the formula in (\ref{eq:ecd_binary}), we are able to determine whether a binary classification model is ``safe'' to use, based on its calibration score, as described in the following section. 

\subsection{ECD Score Interpretation for single datum}

\begin{figure}[ht]
\vskip 0.2in
\begin{center}
\centerline{\includegraphics[width=\columnwidth]{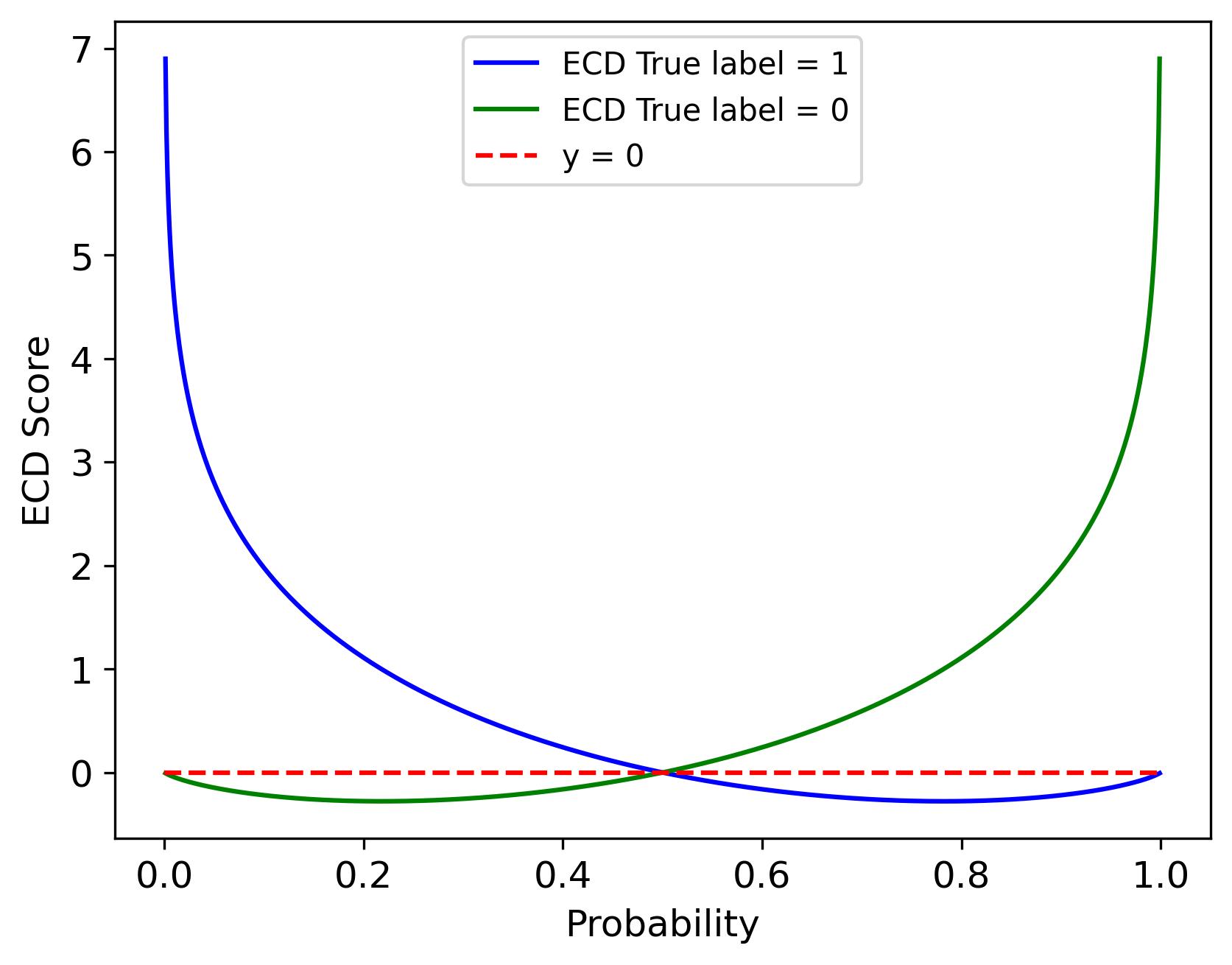}}
\caption{Range of ECD scores for probabilities between 0 and 1.}
\label{fig:ecd_score_range}
\end{center}
\vskip -0.2in
\end{figure}

Unlike the metrics discussed previously, ECD does not have an upper bound but has a lower bound for binary classifiers of approximately -0.2785 for a single data point. Positive scores indicate over-confidence, while negative scores show under-confidence. Figure \ref{fig:ecd_score_range} shows the range of ECD scores for probabilities between 0.0001 and 0.9999. Note how the ECD score for an over-confident prediction is penalised much more than that of an under-confident prediction. It is recommended that a threshold value is used to determine the level of safety that users wish. A score of 0 is given for both perfect calibration, when the estimated probability is unity and the predicted class is correct, and perfect uncertainty, when the estimated probability is 0.5, regardless of the true class. Generally, a score close to 0 should be interpreted as a ``safe'' calibration score, rather than a perfect calibration score. If a user wishes to differentiate between a well-calibrated model and an uncertain model, it is recommended that they use the same binning strategy as ECE and look at the individual bin values. This will give additional insight into whether certain probability ranges are more over-confident than others. Due to the nature of the equation, which aggregates scores for each individual prediction and true label, binning strategies do not change the final ECD score. 
\section{Results \& Analysis}
\label{sec:results}
In this section, we compute the ECD metric, as well as the ECE and ESCE metrics for comparison, for both simulated and real datasets. The highlight of this section will be showcasing how ECD identifies ``safe'' calibrated models, whereas other metrics only determine whether a model is well calibrated or not.

\subsection{Application to Simulated Data}

\subsubsection{Experiment Setup}

To simulate probabilities of the required distribution, we first sample an unscaled log-odds value $u'_{LO}$ from a uniform distribution (\ref{eq:uniform}).
\begin{equation}
    \label{eq:uniform}
    u'_{LO} \sim \text{Uniform}(-10, 10)
\end{equation}
This uniform distribution is scaled by a weight $W$ in (\ref{eq:weight}) to achieve different levels of sharpness in output probabilities. The weighting parameter, $W$, is set to 0.5, which results in a good distribution across all bins, with the majority of probabilities concentrated near 1 and 0.
\begin{equation}
    \label{eq:weight}
    u_{LO} = W \cdot u'_{LO}
\end{equation}
The true probability $\tilde{p}$ of each data point is computed using the standard logistic function in (\ref{eq:logistic}).
\begin{equation}
    \label{eq:logistic}
    \tilde{p} = \frac{1}{1+e^{-u_{LO}}}
\end{equation}
The true labels $L$ of each data point are simulated using a binomial distribution (\ref{eq:binomial}).
\begin{equation}
    \label{eq:binomial}
    L \sim \text{Binomial}(n=1, p=\tilde{p}),
\end{equation}
To simulate miscalibration, the estimated probabilities $\tilde{p}$ are computed by using $u_{LO} + \epsilon$ in (\ref{eq:logistic}) in place of $u_{LO}$, where $\epsilon \sim \mathcal{N}(\mu, \sigma^2)$ is a noisy error term. If $\epsilon=0$, the system is well calibrated.

\begin{figure*}[h]
    \centering
    \captionsetup[subfigure]{}
    \begin{subfigure}[t]{0.32\textwidth}
        \centering
        \includegraphics[width=\textwidth]{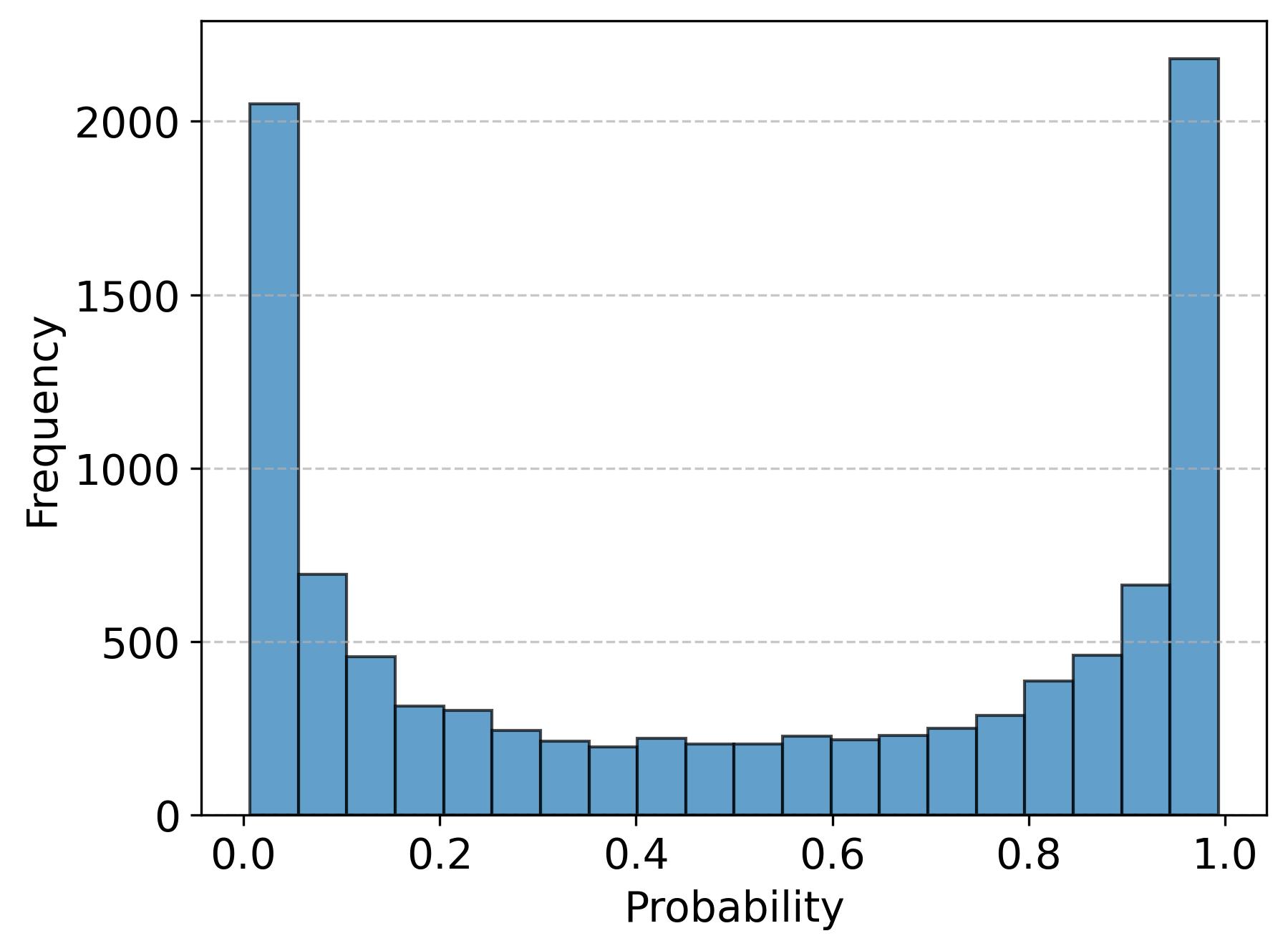}
        \caption{$\epsilon = 0$.}
        \label{fig:hist1}
    \end{subfigure}
    \hfill
    \begin{subfigure}[t]{0.32\textwidth}
        \centering
        \includegraphics[width=\textwidth]{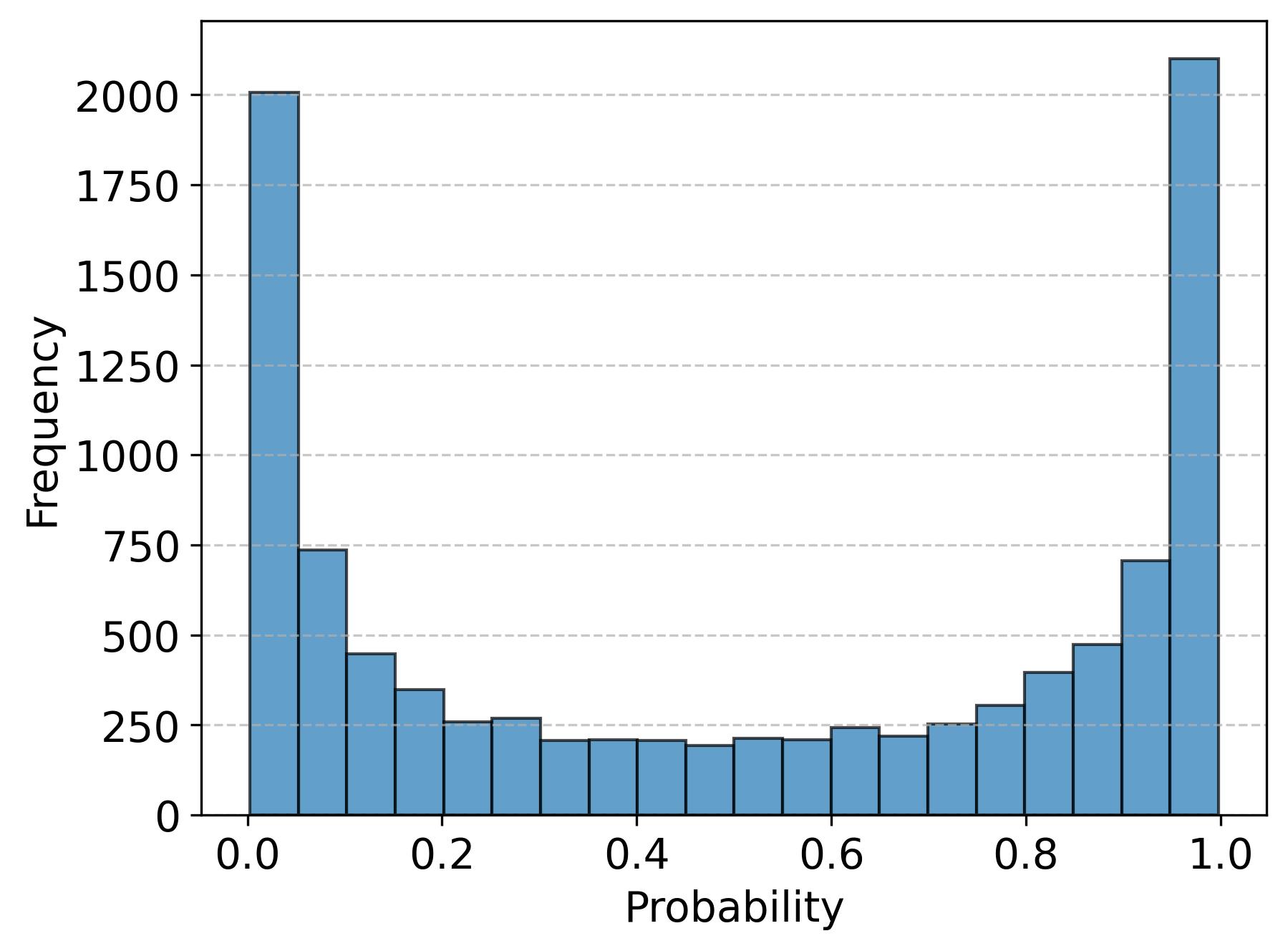}
        \caption{$\epsilon \sim \mathcal{N}(0, 0.5^2)$.}
        \label{fig:hist2}
    \end{subfigure}
    \hfill
    \begin{subfigure}[t]{0.32\textwidth}
        \centering
        \includegraphics[width=\textwidth]{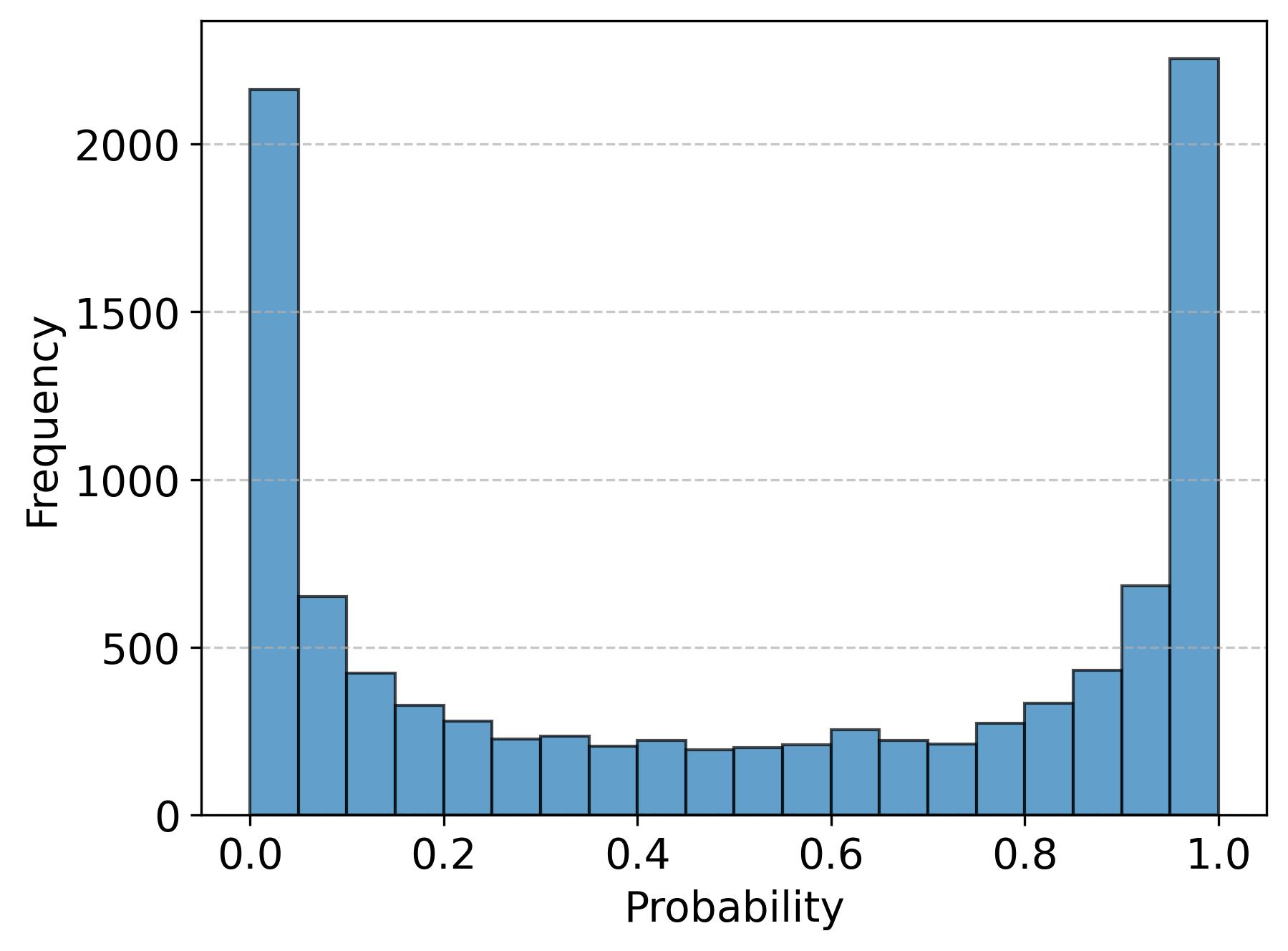}
        \caption{$\epsilon \sim \mathcal{N}(0, 2^2)$.}
        \label{fig:hist3}
    \end{subfigure}
    \caption{Histograms of simulated data after different noise $\epsilon$ added.}
\end{figure*}

For the following tests, 10,000 probabilities were simulated. We perform three different tests, each with a different value of $\epsilon$ to generate a dataset. The first set keeps $\epsilon$ at 0, to demonstrate how each metric treats a well-calibrated model. The second and third sets are generated from a Gaussian distribution, with $\mu$ set to 0 and $\sigma \in \{0.5, 2\}$. Histograms showing the probabilities with these values of $\epsilon$ can be seen in figures \ref{fig:hist1}, \ref{fig:hist2}, and \ref{fig:hist3}, respectively.

\begin{figure*}[h]
    \centering
    \captionsetup[subfigure]{}
    \label{fig:rel_sim}
    \begin{subfigure}[t]{0.32\textwidth}
        \centering
        \includegraphics[width=\textwidth]{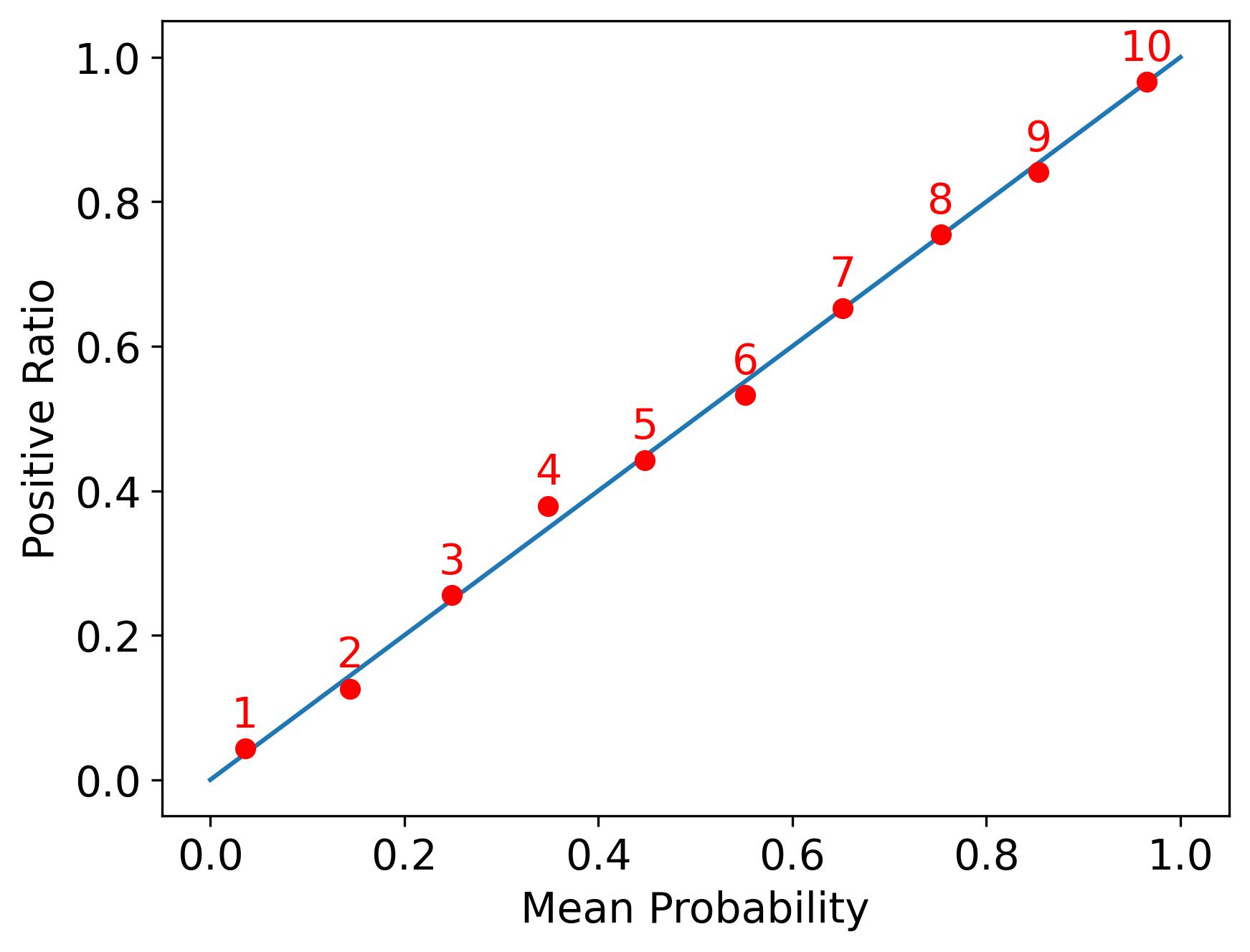}
        \caption{$\epsilon = 0$.}
        \label{fig:rel1_sim}
    \end{subfigure}
    \hfill
    \begin{subfigure}[t]{0.32\textwidth}
        \centering
        \includegraphics[width=\textwidth]{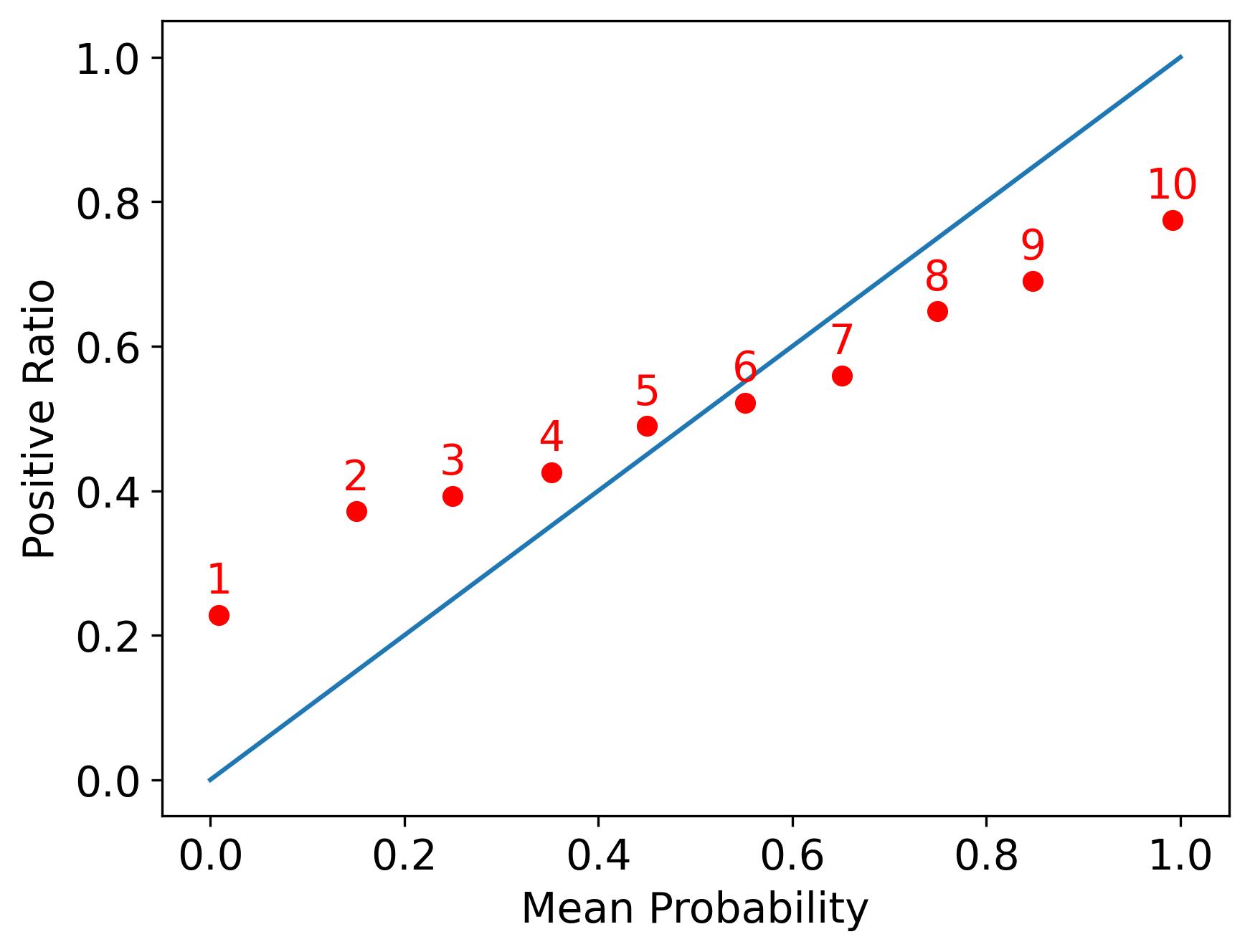}
        \caption{$\epsilon \sim \mathcal{N}(0, 0.5^2)$.}
        \label{fig:rel2_sim}
    \end{subfigure}
    \hfill
    \begin{subfigure}[t]{0.32\textwidth}
        \centering
        \includegraphics[width=\textwidth]{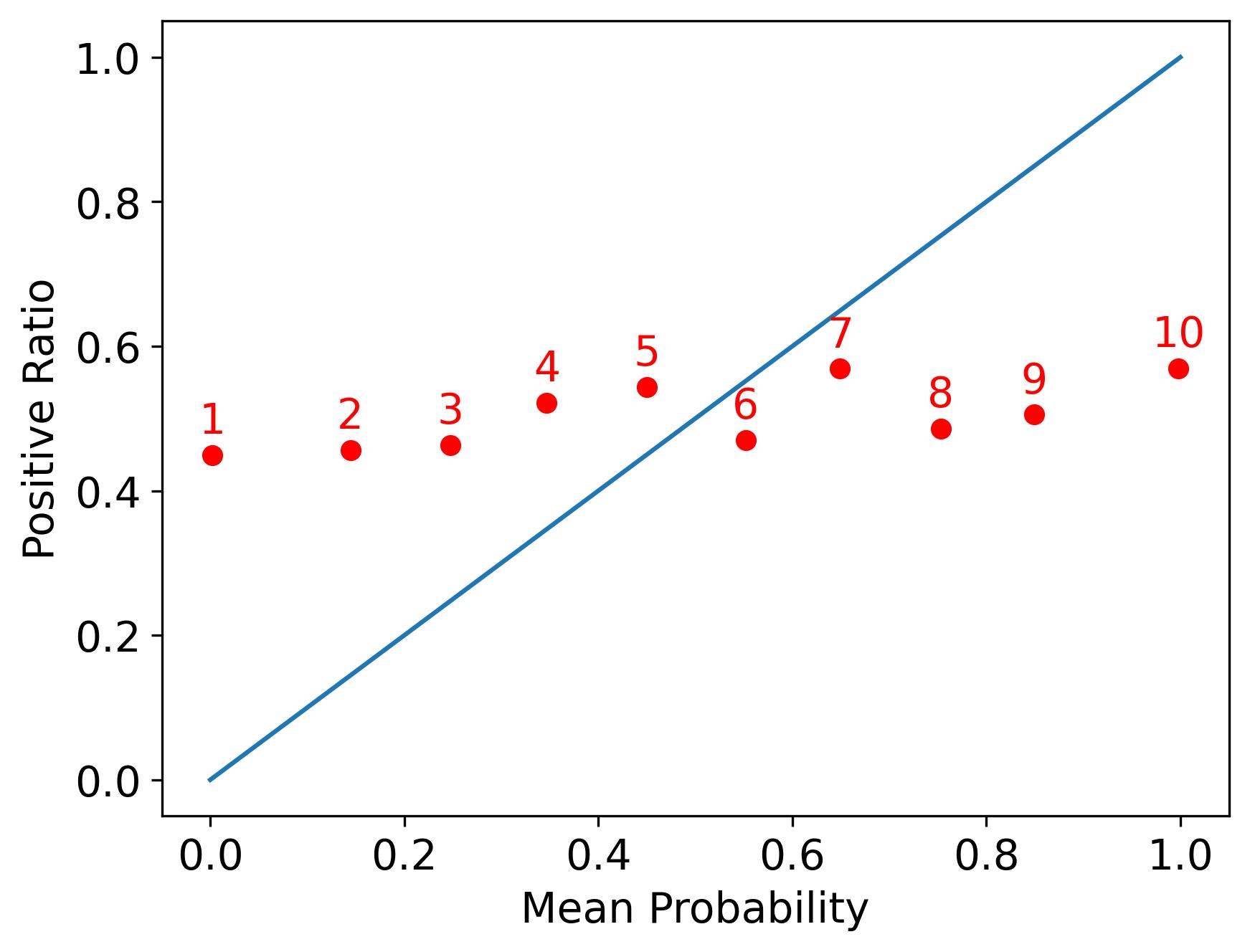}
        \caption{$\epsilon \sim \mathcal{N}(0, 2^2)$.}
        \label{fig:rel3_sim}
    \end{subfigure}
    \caption{Reliability diagrams of simulated data with different noise $\epsilon$ added. Blue line represents $y = x$. Red dots are numbered bins.}
\end{figure*}

We present the scores of the ECD, ECE, and ESCE metrics in Table \ref{tab:table1}. The scores presented are unweighted, with a weighted sum for each metric in the final row. Figures \ref{fig:rel1_sim}, \ref{fig:rel2_sim}, and \ref{fig:rel3_sim} show the reliability diagrams for each set of $\epsilon$ noise values.

\begin{table*}[h]
    \centering
    \caption{Unweighted Simulated Data scores with varying levels of noise $\epsilon$.}
    \begin{tabular}{|c|c|c|c|c|c|c|c|c|c|c|}
        \hline
        \multicolumn{2}{|c|}{} & \multicolumn{3}{|c|}{$\epsilon = 0$} & \multicolumn{3}{|c|}{$\epsilon \sim \mathcal{N}(0, 0.5^2)$} & \multicolumn{3}{|c|}{$\epsilon \sim \mathcal{N}(0, 2^2)$}\\
        \hline
        \textbf{Threshold} & \textbf{Bin} & \textbf{ECE} & \textbf{ESCE} & \textbf{ECD} & \textbf{ECE} & \textbf{ESCE} & \textbf{ECD} & \textbf{ECE} & \textbf{ESCE} & \textbf{ECD} \\
        \hline
        \textbf{0 $\leq$ p $<$ 0.1} & \textbf{1} & 0.0074 & 0.0074 & 0.0231 & 0.2194 & 0.2194 & 2.1406 & 0.4473 & 0.4473 & 4.995\\
        \hline
        \textbf{0.1 $\leq$ p $<$ 0.2}& \textbf{2} & 0.0181 & -0.0181 & -0.0334 & 0.2214 & 0.2214 & 0.3860 & 0.3115 & 0.3115 & 0.5701\\
        \hline
        \textbf{0.2 $\leq$ p $<$ 0.3}& \textbf{3} & 0.0066 & 0.0066 & 0.0069 & 0.1432 & 0.1432 & 0.1582 & 0.2158 & 0.2158 & 0.2529\\
        \hline
        \textbf{0.3 $\leq$ p $<$ 0.4}& \textbf{4} & 0.0306 & 0.0306 & 0.0153 & 0.0743 & 0.0743 & 0.0488 & 0.1749 & 0.1749 & 0.1148\\
        \hline
        \textbf{0.4 $\leq$ p $<$ 0.5}& \textbf{5} & 0.0059 & -0.0059 & 0.0019 & 0.0399 & 0.0399 & 0.0099 & 0.0936 & 0.0936 & 0.0264\\
        \hline
        \textbf{0.5 $\leq$ p $<$ 0.6}& \textbf{6} & 0.0193 & -0.0193 & 0.0036 & 0.0294 & -0.0294 & 0.0045 & 0.0817 & -0.0817 & 0.0246\\
        \hline
        \textbf{0.6 $\leq$ p $<$ 0.7}& \textbf{7} & 0.0011 & 0.0011 & -0.0057 & 0.0911 & -0.0911 & 0.0597 & 0.0803 & -0.0803 & 0.0534\\
        \hline
        \textbf{0.7 $\leq$ p $<$ 0.8}& \textbf{8} & 0.0015 & 0.0015 & 0.0012 & 0.1010 & -0.1010 & 0.1143 & 0.2672 & -0.2672 & 0.3034\\
        \hline
        \textbf{0.8 $\leq$ p $<$ 0.9}& \textbf{9} & 0.0131 & -0.0131 & 0.0242 & 0.1574 & -0.1574 & 0.2771 & 0.3439 & -0.3439 & 0.6044\\
        \hline
        \textbf{0.9 $\leq$ p $\leq$ 1.0}& \textbf{10} & 0.0007 & 0.0007 & -0.0037 & 0.2171 & -0.2171 & 2.1415 & 0.4287 & -0.4287 & 4.8069\\
        \hline
        \hline
        \multicolumn{2}{|c|}{\textbf{Weighted Sum}} & \textbf{0.0077} & \textbf{0.0003} & \textbf{0.0057} & \textbf{0.1702} & \textbf{0.0035} & \textbf{1.2901} & \textbf{0.4042} & \textbf{0.0064} & \textbf{4.2405} \\
        \hline
    \end{tabular}
    \label{tab:table1}
\end{table*}

\begin{table*}[h]
    \centering
    \caption{Unweighted Real Data scores for models BERTimbau \cite{bert_portugal}, ResNet18 \cite{resnet}, and RoBERTa \cite{sensitive}.}
    \begin{tabular}{|c|c|c|c|c|c|c|c|c|c|c|}
        \hline
        \multicolumn{2}{|c|}{} & \multicolumn{3}{|c|}{BERTimbau} & \multicolumn{3}{|c|}{ResNet18} & \multicolumn{3}{|c|}{RoBERTa}\\
        \hline
        \textbf{Threshold} & \textbf{Bin} & \textbf{ECE} & \textbf{ESCE} & \textbf{ECD} & \textbf{ECE} & \textbf{ESCE} & \textbf{ECD} & \textbf{ECE} & \textbf{ESCE} & \textbf{ECD} \\
        \hline
        \textbf{0 $\leq$ p $<$ 0.1} & \textbf{1} & 0.0504 & 0.0504 & 0.3927 & 0.0252 & 0.0252 & 0.0640 & 0.0384 & 0.0384 & 0.1407 \\
        \hline
        \textbf{0.1 $\leq$ p $<$ 0.2} & \textbf{2} & 0.2234 & 0.2234 & 0.3900 & N/A & N/A & N/A & 0.4531 & 0.4531 & 0.7836 \\
        \hline
        \textbf{0.2 $\leq$ p $<$ 0.3} & \textbf{3} & 0.1548 & 0.1548 & 0.1718 & N/A & N/A & N/A & 0.0882 & -0.0882 & -0.0476 \\
        \hline
        \textbf{0.3 $\leq$ p $<$ 0.4} & \textbf{4} & 0.2265 & 0.2265 & 0.1417 & N/A & N/A & N/A & 0.0192 & 0.0192 & -0.0006 \\
        \hline
        \textbf{0.4 $\leq$ p $<$ 0.5} & \textbf{5} & 0.0723 & -0.0723 & -0.0213 & N/A & N/A & N/A & 0.2952 & 0.2952 & 0.0408 \\
        \hline
        \textbf{0.5 $\leq$ p $<$ 0.6} & \textbf{6} & 0.0792 & -0.0792 & 0.0019 & N/A & N/A & N/A & 0.5455 & -0.5455 & 0.1008 \\
        \hline
        \textbf{0.6 $\leq$ p $<$ 0.7} & \textbf{7} & 0.1515 & -0.1515 & 0.0952 & 0.3800 & 0.3800 & -0.1860 & 0.0880 & 0.0880 & -0.0239 \\
        \hline
        \textbf{0.7 $\leq$ p $<$ 0.8} & \textbf{8} & 0.2721 & -0.2721 & 0.3014 & 0.2132 & 0.2132 & -0.2784 & 0.2696 & -0.2696 & 0.2623 \\
        \hline
        \textbf{0.8 $\leq$ p $<$ 0.9} & \textbf{9} & 0.4006 & -0.4006 & 0.7424 & N/A & N/A & N/A & 0.0266 & -0.0266 & 0.0105 \\
        \hline
        \textbf{0.9 $\leq$ p $\leq$ 1.0} & \textbf{10} & 0.2953 & -0.2953 & 1.5816 & 0.0127 & -0.0127 & 0.0649 & 0.0439 & -0.0439 & 0.1799 \\
        \hline
        \hline
        \multicolumn{2}{|c|}{\textbf{Weighted Sum}} & \textbf{0.0772} & \textbf{0.0203} & \textbf{0.4767} & \textbf{0.0231} & \textbf{0.0109} & \textbf{0.0602} & \textbf{0.0456} & \textbf{-0.0049} & \textbf{0.1629} \\
        \hline
    \end{tabular}
    \label{tab:table2}
\end{table*}

\subsubsection{Analysis}

For the first test, we set $\epsilon$ to 0 which results in the distribution of probabilities over the 10 bins as seen in figure \ref{fig:hist1}. In the corresponding column of Table \ref{tab:table1}, we can see that the weighted sums for each metric are very close to 0. This indicates that the model would be both well-calibrated and ``safe'' to use. Looking at figure \ref{fig:rel1_sim}, we can see that the reliability diagram also showcases the well-calibrated nature of the probabilities. 

In the second test, we set $\epsilon$ to sample values from a Gaussian distribution using a mean $\mu$ of 0 and standard deviation $\sigma$ of 0.5. The results for each metric can be seen in corresponding column of Table \ref{tab:table1}. The low ESCE metric indicates the probabilities are globally calibrated but the ECE shows that there is some local miscalibration. This is backed up by the reliability diagram in figure \ref{fig:rel2_sim}.
The ECD metric highlights a considerable amount of over-confidence in the first and last bins. Figure \ref{fig:hist2} shows how these two bins are the most populated and carry the most weight. While the ECE and ESCE scores may give the impression that these probabilities are either only mildly miscalibrated or globally well-calibrated, the ECD score shows that, in fact, they are not ``safe''. This case shows that, while a metric such as ECE shows generally good calibration, it does not always mean that the probabilities are ``safe'' to use. 

In the third and final test, we set $\epsilon$ to sample values from a Gaussian distribution with $\mu$ set to 0 and $\sigma$ set to 2. While the distribution of values in figure \ref{fig:hist3} looks similar to the previous two, the reliability diagram in figure \ref{fig:rel3_sim} shows a large amount of miscalibration. The corresponding column in Table \ref{tab:table1} shows the ECE metric highlighting the miscalibration. The ECD scores show that there are high levels of overconfidence in the outer bins. In this case, the ECE and ECD agree on miscalibration, while the ESCE weighted sum has been significantly lowered due to the contrasting signs.

\subsection{Application to Real Data}
\subsubsection{Experiment Setup}

We conducted experiments similar to the above using real data on pre-trained binary models with good accuracy scores. All models were obtained using the Hugging Face transformers library \cite{huggingface}. The first model is a fine-tuned binary version of the BERTimbau model \cite{bert_portugal} that classifies Portuguese sentences as hate speech or not, using a BERT language model \cite{bert}. The second model is a fine-tuned version of ResNet18 \cite{resnet} that identifies whether an image is a cat or a dog using binary classification. Lastly, we used a RoBERTa model that classifies online comments into a sensitive or non-sensitive class \cite{sensitive}. 

\begin{figure*}[!htb]
    \centering
    \captionsetup[subfigure]{}
    \begin{subfigure}[t]{0.32\textwidth}
        \centering
        \includegraphics[width=\textwidth]{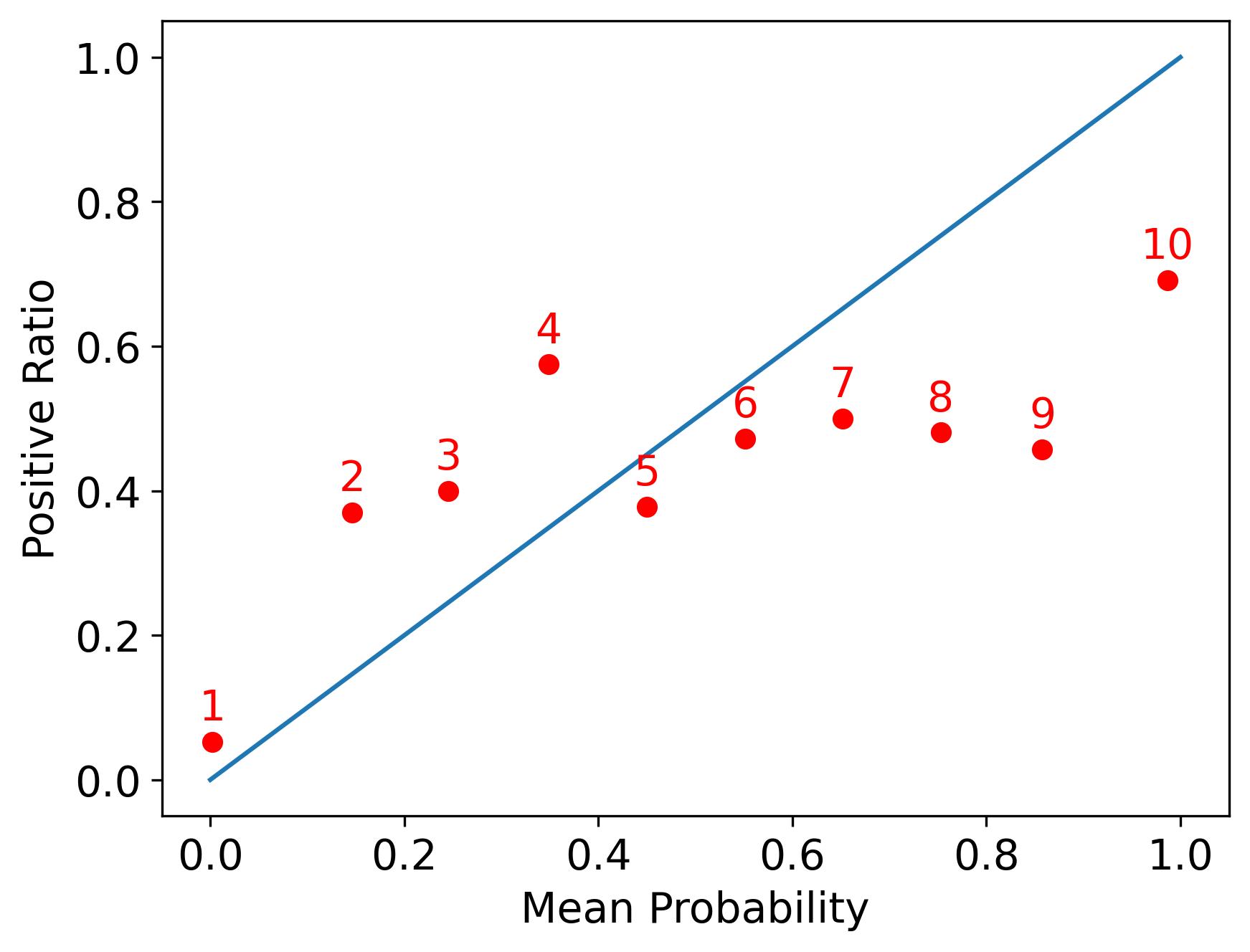}
        \caption{BERTimbau}
        \label{fig:model1}
    \end{subfigure}
    \hfill
    \begin{subfigure}[t]{0.32\textwidth}
        \centering
        \includegraphics[width=\textwidth]{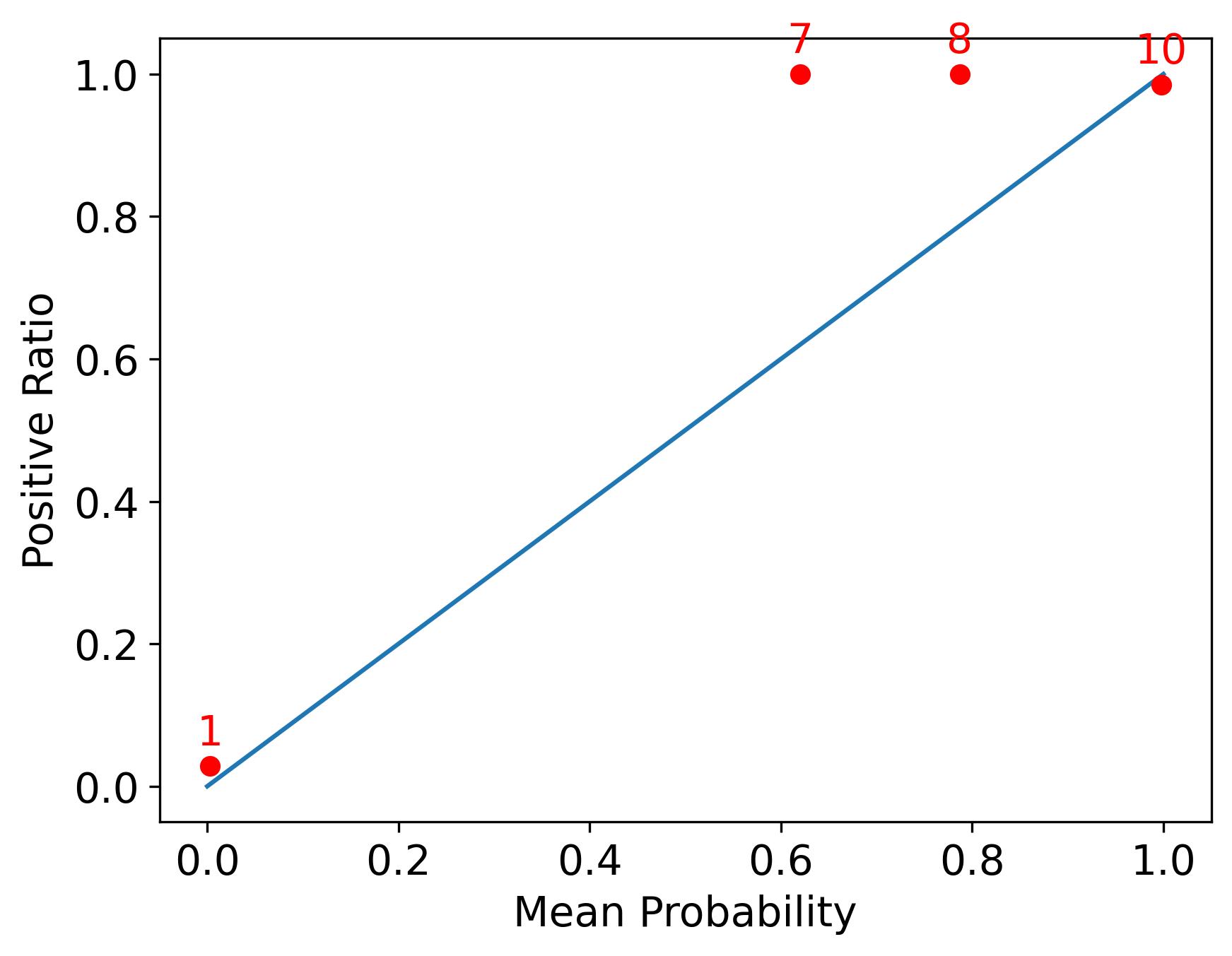}
        \caption{ResNet18}
        \label{fig:model2}
    \end{subfigure}
    \hfill
    \begin{subfigure}[t]{0.32\textwidth}
        \centering
        \includegraphics[width=\textwidth]{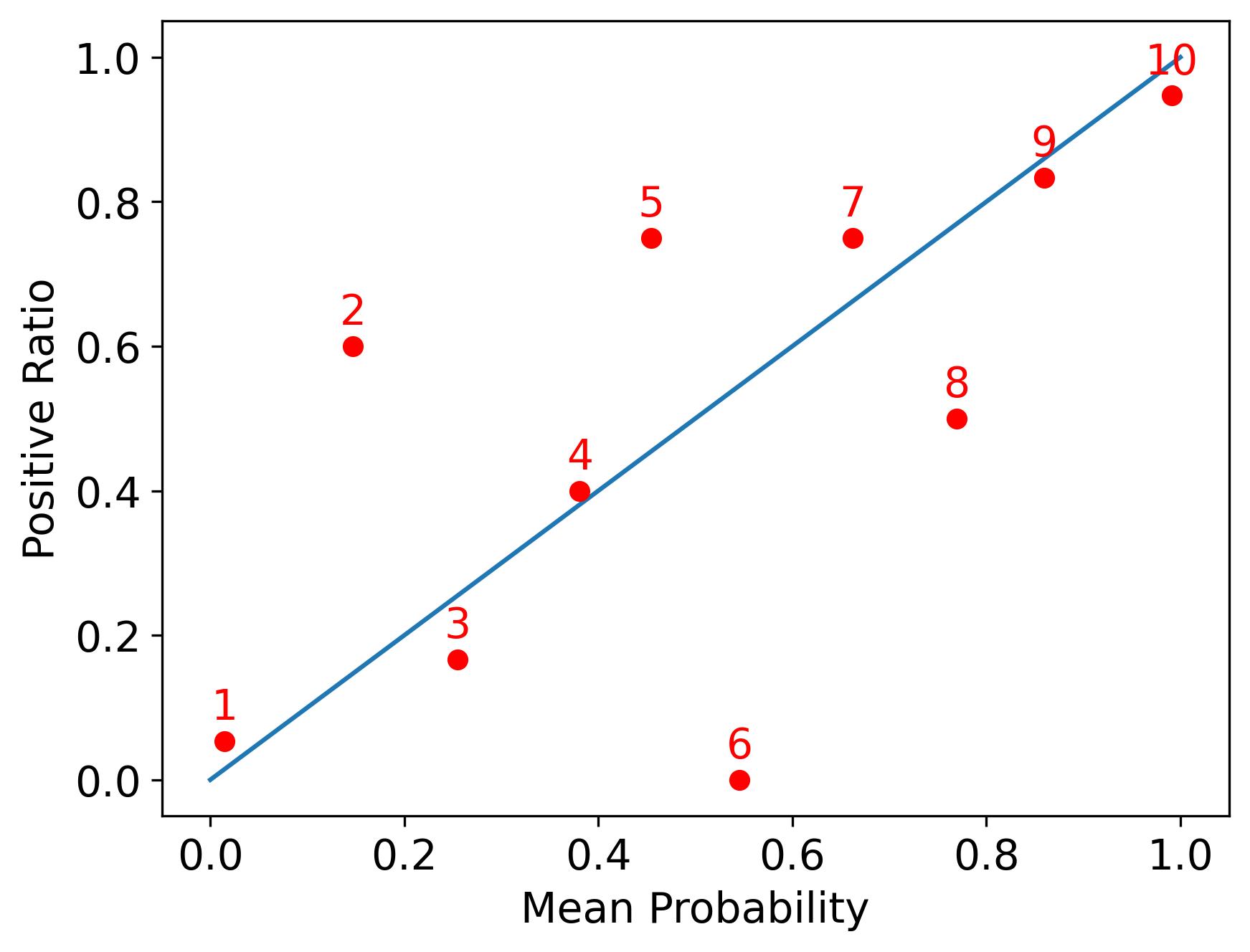}
        \caption{RoBERTa}
        \label{fig:model3}
    \end{subfigure}
    \label{fig:models}
    \caption{Reliability diagrams of (a) BERTimbau, (b) ResNet18, and (c) RoBERTa.}
\end{figure*}

For each model, unseen data was used and the output probabilities were split into ten bins. The ECE, ESCE and ECD metrics were calculated and the unweighted bin values can be seen in Table \ref{tab:table2}. The reliability diagrams for each model are shown in figures \ref{fig:model1}, \ref{fig:model2}, and \ref{fig:model3}. 

\subsubsection{Analysis}

This section discusses the scores obtained in each bin, as shown in Table \ref{tab:table2}. 

In the BERTimbau model, the ECE and ESCE scores suggest that the model is very well-calibrated on the whole, with final scores of 0.0772 and 0.0203, respectively. However, the ECD metric shows a different perspective, and highlights that Bins 1 and 10 could be a cause for concern, due to some predictions being highly confident in the incorrect class. If this over-confidence in the incorrect class is a cause for concern for the user, the ECD provides additional insight which is not present in the ECE and ESCE scores. 

Due to the confidence in the predictions, not all of the bins for ResNet18 are populated. The majority of predictions are in the first and last bins. This time, the ECE and ESCE metrics indicate a very well-calibrated model, and the ECD score agrees. Therefore, it can be concluded that this model is well-calibrated and safe to use. 

The ECE score for the RoBERTa model indicates that it is better calibrated than BERTimbau, but worse than ResNet18. On the other hand, the ESCE scores suggest that the RoBERTa model is better calibrated than the other two models. The majority of values are once again in the first and last bins, indicating confidence of the model's predictions. This is a similar situation to the BERTimbau model, where the ECE and ESCE metrics indicate a well-calibrated metric, while the ECD indicates that there is some possible unsafe actions being made in bins 1 and 10, although these are not as bad as the other example. 

Overall, these results show that even if a model is apparently considered well-calibrated in terms of ECE and ESCE scores, that does not necessarily indicate that they are ``safe'' to use. By combining all metrics together as seen in ResNet18, we are able to deduce that the model is well-calibrated and ``safe'' to use. Therefore, the ECD metric can introduce another perspective to calibration which is not always considered within the ECE and ESCE metrics. 

\section{Conclusion and Future Work}
\label{sec:conclusion}
We have introduced a novel calibration metric, named the Entropic Calibration Difference (ECD) due to its consideration of the entropy of probabilities. The metric is influenced by the Normalised Estimation Error Squared (NEES) metric, which is used to determine the consistency of a state estimator within the target-tracking field of research. The ECD metric is equivalent to applying a generalised version of NEES to the ML problem domain. This new metric also brings a new perspective to the probability calibration literature, namely the concept of ``safe'' calibration, which is commonly found in the target tracking literature. We define ``safe'' calibration as a metric that prefers under-confidence to over-confidence, due to the belief that an under-confident score in the correct class is safer than an overconfident score in the incorrect class. Therefore, over-confidence is penalised more than under-confidence, rather than equal penalties, which are present in other metrics. 

In terms of future directions, we believe that the ECD metric could be applied to other problems, and the concept of ``safe'' calibration could be expanded upon further, and even implemented into existing popular metrics. 

\bibliography{references}
\bibliographystyle{IEEEtran}

\end{document}